\pdfoutput=1

\documentclass[11pt,a4paper]{article}

\usepackage[T1]{fontenc}
\usepackage{lmodern}
\input{glyphtounicode}
\usepackage[utf8]{inputenc}
\usepackage[margin=1in]{geometry}
\usepackage{amssymb}
\usepackage{amsmath}
\usepackage{graphicx}
\usepackage{booktabs}
\usepackage{array}
\usepackage{url}
\usepackage{xurl}
\usepackage{authblk}
\usepackage{caption}
\usepackage{setspace}
\usepackage{float}
\usepackage{tikz}
\usetikzlibrary{positioning,arrows.meta}

\title{Real-time pedestrian attribute recognition with YOLOv8 and ResNet18}

\author[1]{Houssam El Mir\thanks{Corresponding author. E-mail addresses: l202126100102@zjut.edu.cn; houssamelmir2020@gmail.com}}
\affil[1]{College of Computer Science and Technology, Zhejiang University of Technology, Hangzhou, 310023, Zhejiang, China}
\date{}

\begin{document}

\maketitle

\begin{abstract}
Pedestrian attribute recognition (PAR) predicts semantic labels from pedestrian images and has applications in video retrieval and human-centred graphics. This paper evaluates ResNet18 classifiers for gender, four-class age-group prediction, and multi-attribute recognition using PETA and PA-100K dataset crops. On the PETA identity-split test set, gender accuracy is $0.8087 \pm 0.0119$; four-class age-group accuracy is $0.5481 \pm 0.0263$ with macro-F1 of $0.3914 \pm 0.0361$; and PETA-35 mean accuracy-based mA is $0.7249 \pm 0.0049$, with mean average precision (mAP) of $0.5284 \pm 0.0069$. The YOLOv8n detector finds a person in 79.02\% of PETA test crops, but detection-to-attribute accuracy was not measured end to end. On the eight targets shared by PETA and PA-100K, combined training yielded lower mA than PETA-only training, with the drop consistent across seeds. Batch-one stage latency was measured on an NVIDIA RTX 5060 GPU.
\end{abstract}


\section{Introduction}

Pedestrian attribute recognition (PAR) predicts semantic attributes, such as clothing type, carried objects, gender, and apparent age, from pedestrian images. PAR is relevant to surveillance, transportation analytics, human-computer interaction, and human-centred graphics applications such as virtual try-on and avatar generation. In video analysis, these attributes convert unstructured frames into searchable metadata~\cite{li2016,deng2014}.

Deep learning has substantially improved visual recognition since AlexNet~\cite{krizhevsky2012}. Architectures including VGG~\cite{simonyan2014}, GoogLeNet~\cite{szegedy2015}, ResNet~\cite{he2016}, DenseNet~\cite{huang2017}, and EfficientNet~\cite{tan2019} provide strong feature extractors for PAR. However, practical PAR remains difficult because pedestrian crops vary in scale, viewpoint, illumination, and occlusion; labels are multi-valued; detection errors propagate to recognition; and many attributes are highly imbalanced~\cite{sarafianos2016,li2015}.

This work evaluates recognition accuracy on PETA and PA-100K dataset crops, using an identity-disjoint PETA test split for the reported PETA results. It also evaluates the YOLOv8n detector separately: a person is detected in 79.02\% of PETA test crops. Detection-to-attribute accuracy was not measured end to end, and detector-generated crops were not used to produce the recognition accuracy results. The attribute experiments follow the PETA-35 protocol; the PA-100K mapping covers eight shared target attributes. The work therefore reports crop-based recognition results alongside a separate detector assessment, rather than an end-to-end accuracy evaluation.

This paper presents an integrated YOLOv8n--ResNet18 framework with the following contributions:
\begin{itemize}
    \item A two-stage detection and recognition pipeline combining YOLOv8n with ResNet18-based task heads.
    \item Crop-based evaluation of gender, four-class age-group, and PETA-35 attribute recognition.
    \item A PA-100K-to-PETA mapping for the eight shared target attributes.
    \item Separate evaluation of person detection and per-stage latency on an NVIDIA RTX 5060 GPU, without an end-to-end detection-to-attribute accuracy claim.
\end{itemize}

\section{Related Work}

\subsection{Pedestrian Detection}
Early detection used HOG with SVM classifiers~\cite{dalal2005}. Deep learning methods like Faster R-CNN~\cite{ren2015} improved accuracy but were computationally expensive. YOLO~\cite{redmon2016} revolutionized real-time detection. YOLOv8~\cite{ultralytics2023} introduces anchor-free detection, ideal for surveillance.

For a practical PAR system, detection quality is important because every downstream prediction depends on the crop produced by the detector. Missed detections remove pedestrians from the recognition stage, while loose or inaccurate bounding boxes may include background clutter that reduces attribute accuracy. The use of a lightweight YOLOv8n detector is therefore motivated by the need to balance localization quality, throughput, and ease of deployment in video streams.

\subsection{Attribute Recognition}
Early PAR approaches trained separate classifiers per attribute, ignoring correlations among clothing, carried objects, and demographic cues. Deep learning enables shared feature extraction and joint multi-label learning. Recent work explores attention mechanisms~\cite{liu2017}, multi-task learning~\cite{wang2018}, imbalanced classification~\cite{sarafianos2016}, and multi-attribute learning~\cite{li2015}. In computer graphics, attribute recognition can provide semantic inputs for style transfer, pose-aware rendering, and human-centric scene analysis.

More recent work has continued to develop Transformer-based PAR models.
Zheng et al.~\cite{zheng2023} suppress the most discriminative feature
regions to force attention toward previously neglected attribute-relevant
areas. Jin et al.~\cite{jin2024} reformulate attribute prediction as
sequence generation with a Transformer decoder. Wang et al.~\cite{wang2024}
study Mamba-based alternatives to Transformer backbones for PAR, and Tan
et al.~\cite{tan2024} explore relation modelling between attributes within
a vision Transformer. These approaches target architectural and
representational improvements and do not evaluate an identity-disjoint
split or a detection-to-attribute pipeline, which is the focus of the
present work.

Modern PAR models usually share visual features across attributes because many labels are visually correlated. For example, lower-body clothing type, clothing colour, and carried-object labels can often be inferred from overlapping image regions. This motivates the ResNet18-based design used in this work, where task-specific models handle gender classification, age-group classification, and multi-label attribute prediction.

\subsection{Datasets}
Key datasets include PETA~\cite{deng2014}, RAP~\cite{li2016}, and PA-100K~\cite{liu2017}. Related person re-identification datasets, such as Market-1501 and DukeMTMC, have also been used with attribute or identity supervision~\cite{lin2019,li2016reid}. The experiments use the PETA-35 attribute protocol; the PA-100K mapping is limited to the eight targets shared with PETA in the combined experiment.

The datasets use different annotation policies, so the combined experiment maps only the eight shared targets and evaluates them on the PETA identity-split test set. The comparison does not assume that unshared labels have exact counterparts.

\section{Methodology}

\subsection{System Architecture}
The system follows the two-stage pipeline in Fig.~\ref{fig:pipeline}. YOLOv8n~\cite{ultralytics2023} detects people in input frames. Recognition accuracy, however, is evaluated separately on dataset-provided PETA and PA-100K crops; detector-generated crops are not used for those accuracy measurements. ResNet18-based classifiers process the crops~\cite{he2016}.

\begin{figure}[H]
\centering
\begin{tikzpicture}[
    node distance=0.75cm and 0.9cm,
    stage/.style={draw, rounded corners, align=center, minimum width=4.0cm, minimum height=0.8cm},
    head/.style={draw, rounded corners, align=center, minimum width=2.8cm, minimum height=0.75cm},
    flow/.style={-{Stealth}, thick}
]
\node[stage] (input) {Input frame};
\node[stage, below=of input] (detector) {YOLOv8n detector};
\node[stage, below=of detector] (crop) {Crop and resize\\$224\times224$};
\node[stage, below=of crop] (backbone) {ResNet18};
\node[head, below left=0.85cm and 1.25cm of backbone] (gender) {Gender head};
\node[head, below=0.85cm of backbone] (age) {Age head};
\node[head, below right=0.85cm and 1.25cm of backbone] (attributes) {35-attribute head};
\node[stage, below=1.0cm of age] (output) {Annotated output};
\draw[flow] (input) -- (detector);
\draw[flow] (detector) -- (crop);
\draw[flow] (crop) -- (backbone);
\draw[flow] (backbone) -- (gender);
\draw[flow] (backbone) -- (age);
\draw[flow] (backbone) -- (attributes);
\draw[flow] (gender) |- (output);
\draw[flow] (age) -- (output);
\draw[flow] (attributes) |- (output);
\end{tikzpicture}
\caption{Two-stage PAR pipeline from frame input to annotated pedestrian output.}
\label{fig:pipeline}
\end{figure}

\subsection{Implementation Workflow}
The intended workflow passes each input frame to YOLOv8n, crops detected people, resizes and normalizes each crop, and associates the task-specific classifier outputs with the corresponding box. The recognition evaluations reported here instead use dataset-provided crops, so this workflow description should not be read as an end-to-end evaluation protocol.

\subsection{Dataset Preparation}
The attribute experiment uses PETA-35~\cite{deng2014}; PA-100K~\cite{liu2017} is included in the combined training comparison only through the eight mapped shared targets. Images are resized to $224\times224$ and normalized with ImageNet statistics: mean $=[0.485, 0.456, 0.406]$ and standard deviation $=[0.229, 0.224, 0.225]$~\cite{krizhevsky2012,he2016}.

\textbf{Combined Strategy:} The attribute experiments use the PETA-35 protocol. For the combined comparison, PA-100K labels are mapped only to the eight target attributes shared with PETA; no broader mapping to PETA's full original attribute set is claimed. Recognition metrics are evaluated on dataset-provided pedestrian crops.

Table~\ref{tab:mapping} lists the eight mappings that the combined comparison uses. Seven are exact, in the sense that the PA-100K label and the PETA-35 label denote the same concept with the same polarity. The eighth, \texttt{Female} to \texttt{personalMale}, is an inversion: PA-100K sets \texttt{Female}=1 for a female pedestrian while PETA sets \texttt{personalMale}=1 for a male pedestrian, so the target is computed as one minus the source label.

\begin{table}[H]
\centering
\caption{The eight PA-100K to PETA-35 mappings used in the combined comparison. Seven are exact and one is inverted.}
\label{tab:mapping}
\small
\renewcommand{\arraystretch}{1.12}
\setlength{\tabcolsep}{4pt}
\begin{tabular}{ll}
\hline
\textbf{PETA-35 target} & \textbf{PA-100K source} \\
\hline
\texttt{accessoryHat} & \texttt{Hat} \\
\texttt{carryingBackpack} & \texttt{Backpack} \\
\texttt{lowerBodyShorts} & \texttt{Shorts} \\
\texttt{lowerBodyTrousers} & \texttt{Trousers} \\
\texttt{upperBodyLogo} & \texttt{UpperLogo} \\
\texttt{upperBodyPlaid} & \texttt{UpperPlaid} \\
\texttt{upperBodyShortSleeve} & \texttt{ShortSleeve} \\
\texttt{personalMale} & \texttt{Female} (inverted) \\
\hline
\end{tabular}
\end{table}

For a PA-100K image, the other 27 PETA-35 attributes are treated as missing and masked out of the loss, not as negatives: those attributes are not supervised by the PA-100K data, so they contribute nothing to the gradient for that sample rather than being taught as absent.

\subsection{Model Architecture}
\textbf{YOLOv8n:} The detector uses COCO-pretrained weights in inference mode. Frames are resized to $640\times640$ pixels, with a confidence threshold of 0.25 and a non-maximum-suppression IoU threshold of 0.45.

\textbf{ResNet18:} Three ImageNet-pretrained ResNet18 networks are fine-tuned for the downstream tasks. The gender model uses two output neurons, the age-group model predicts four classes, and the multi-attribute model uses 35 sigmoid outputs with positive-class weighting. The implementation uses PyTorch~\cite{paszke2019}.

\subsection{Training}
The three task heads were trained with different hyperparameters, and the
PETA-35 attribute head is reported using configuration~C, which was
selected on the validation set: configurations~A, B and C were compared in
an exploration phase in which only validation metrics were computed and
the test set was never loaded, configuration~C was chosen as the best of
the three by validation true mA, and the selected configuration was then
re-run on seeds 0, 1 and~2 to produce the reported test numbers. In every
case the reported checkpoint is the epoch that maximised the validation
metric of that head, and the test set is evaluated exactly once, at the
end, from that checkpoint.

\paragraph{Gender and age-group heads.}
Both heads are ImageNet-pretrained ResNet18 networks with a single linear
classification head, trained with Adam with PyTorch default betas (0.9,
0.999) at learning rate 0.001 for up to 50 epochs with a batch size of 32
and early-stopping patience of 10 epochs; no weight decay was applied.
The gender head has two outputs and uses the cross-entropy loss
without positive-class weighting; its checkpoint is the epoch with the
highest validation accuracy. The age-group head has four outputs and also
uses the cross-entropy loss without positive-class weighting; its
checkpoint is the epoch with the highest validation macro-F1. The training
augmentation for both heads is a random horizontal flip with probability
0.5; no rotation, resized crop or colour jitter is applied, and the
validation and test images receive neither the flip nor any other
augmentation.

\paragraph{PETA-35 attribute head (configuration C).}
The reported attribute model is configuration~C, an ImageNet-pretrained
ResNet18 with 35 sigmoid outputs trained with AdamW with PyTorch default
betas (0.9, 0.999) at learning rate 0.0001 and weight decay 0.01, for up
to 50 epochs with a batch size of 32 and early-stopping patience of 10
epochs. Its loss is a masked binary
cross-entropy with logits,
\begin{equation}
L_{attr,i} = -[w_{pos,i} \cdot y_i \log(\sigma(x_i)) + (1-y_i)\log(1-\sigma(x_i))]
\end{equation}
where $w_{pos,i}$ is the positive-class weight of attribute~$i$ and
$\sigma$ is the logistic sigmoid; this weighted form applies to
configuration~C only, and configuration~B uses the same expression with
$w_{pos,i}=1$. The positive-class weights are computed from the training
split alone as the ratio of negative to positive examples per attribute,
and are clipped to the range $[1, 50]$; the weighting addresses the class
imbalance of the rare attributes~\cite{sarafianos2016}. Its checkpoint is
the epoch with the highest three-epoch moving average of validation true
mA. The training augmentation is a random resized crop with scale in
$[0.7, 1.0]$ to $224\times224$ pixels, a random rotation of $\pm10^\circ$,
colour jitter with brightness, contrast and saturation factors of 0.2 and
a hue factor of 0.1, and a random horizontal flip with probability 0.5;
the validation and test images receive no augmentation. The probabilities
are binarised at a fixed threshold of 0.5.

\paragraph{Positive-class weighting ablation (configuration B versus C).}
Configuration~B and configuration~C share the same optimizer, learning
rate, weight decay, batch size, epoch budget, early-stopping criterion and
augmentation pipeline, and differ only in the positive-class weighting of
the loss: configuration~B sets $w_{pos,i}=1$ for every attribute, while
configuration~C uses the training-split weights described above. Reporting
the two configurations side by side therefore isolates the effect of the
positive-class weighting alone.

For reproducibility, classification metrics are computed on held-out dataset crops using the identity-disjoint PETA test split for PETA evaluations. Detection uses fixed YOLOv8n thresholds, and classification uses the same ImageNet normalization for all tasks. The detector assessment is reported separately and is not combined with the crop-based recognition metrics.

\section{Experiments and Results}

\subsection{Experimental Setup}
Experiments use PETA and PA-100K dataset crops. Gender, age-group, and PETA-35 attribute recognition accuracy is measured on these dataset-provided crops, not on detector-generated crops. The PETA test results use an identity-disjoint split. The detector is evaluated separately, and the study does not measure detection-to-attribute accuracy end to end. Timing measurements are reported for the NVIDIA RTX 5060 GPU.

\subsection{Dataset Splits and Protocols}
PETA gender, age-group, and attribute evaluations use the identity-disjoint test split. The PETA attribute experiment follows the PETA-35 protocol. In the combined experiment, PA-100K labels are mapped to the eight shared targets, and the reported comparison is evaluated on the PETA identity-split test set. Gender classification was also run with a random split; that result is included only to demonstrate the effect of identity leakage and is not treated as the evaluation result.

The identity split is formed by partitioning on the person identifier rather than on the image, so that no pedestrian appears in more than one partition. The partition sizes are given in Table~\ref{tab:split}. No identity is shared between any two of the three partitions, so the overlap is zero for every pair.

To measure how the standard PETA protocol behaves, we also computed the identity overlap of the five official partitions stored in PETA.mat. Every official partition shares identities between its training and test sets, with 1,004 to 1,016 identities in common across the five partitions; these figures are reported only as evidence for the choice of an identity-disjoint split and are not used for any result in this paper.

\begin{table}[H]
\centering
\caption{The PETA identity split, by partition.}
\label{tab:split}
\small
\renewcommand{\arraystretch}{1.12}
\setlength{\tabcolsep}{4pt}
\begin{tabular}{lrr}
\hline
\textbf{Partition} & \textbf{Images} & \textbf{Identities} \\
\hline
Train & 13254 & 6089 \\
Validation & 1805 & 870 \\
Test & 3941 & 1740 \\
\hline
\end{tabular}
\end{table}

A near-duplicate check was run on the image collections to test whether the two datasets share images and whether the PA-100K training and test sets overlap with each other. The check uses ImageNet-pretrained ResNet18 penultimate features and cosine similarity to the nearest neighbour, and it is a report-only check. The maximum nearest-neighbour similarity between PETA test images and the PA-100K training set is 0.9134, and between PA-100K test images and the PETA training set it is 0.9036, both below the 0.95 threshold. Within PA-100K, the maximum nearest-neighbour similarity between a test image and the training set is 0.9563, and 1 query image out of 10000 exceeds the 0.95 threshold. PA-100K provides no person identifiers, so an identity-disjoint split cannot be constructed for it; the official PA-100K training and test partition is used as given.

\subsection{Evaluation Metrics}
Gender and age-group recognition are summarized using test accuracy; age-group recognition is also reported with test macro-F1. PETA-35 attribute performance is reported using mean accuracy-based mA and mean average precision (mAP). Mean accuracy-based mA is the mean across attributes of balanced accuracy, computed for each attribute as $(\mathrm{TPR}+\mathrm{TNR})/2$. mAP is reported separately as the mean average precision across attributes.

\subsection{Comparison with State-of-the-Art Methods}
Tables~\ref{tab:comparison} and~\ref{tab:gender_age_sota} summarize representative approaches in pedestrian attribute recognition, gender classification, and age estimation. This comparison is contextual rather than a strict leaderboard because datasets, annotation vocabularies, crop-generation procedures, and target definitions differ across studies. In the present work, recognition scores are measured on dataset crops, while person detection is assessed separately; end-to-end detection-to-attribute accuracy was not measured.

Many PAR approaches evaluate recognition on cropped pedestrian images and optimize the multi-label classifier. This work also evaluates YOLOv8n detection, but the recognition results do not use detector-generated crops and are not an end-to-end pipeline accuracy measurement. The ResNet18 results should therefore be interpreted as crop-based recognition performance, separately from the detector assessment.

\begin{table}[H]
\centering
\caption{Comparison with representative state-of-the-art PAR methods. Published figures are taken from the original papers and from the survey tables in \cite{jia2020}. Protocol differences mean that the comparison is indicative rather than a strict leaderboard: the published numbers use the standard PETA random split, in which identities appear in both the training and the test partition, whereas this work uses an identity-disjoint split, so the published figures are inflated relative to ours.}
\label{tab:comparison}
\footnotesize
\renewcommand{\arraystretch}{1.12}
\setlength{\tabcolsep}{3pt}
\begin{tabular}{p{0.19\linewidth}lr rrr rr}
\hline
\textbf{Method} & \textbf{Backbone} & \textbf{PETA mA} & \textbf{PETA Accu} & \textbf{PETA F1} & \textbf{PA-100K mA} & \textbf{PA-100K F1} \\
\hline
DeepMAR~\cite{li2015} & CaffeNet & 82.89 & 75.07 & 83.41 & 72.70 & 81.32 \\
HydraPlus-Net~\cite{liu2017} & InceptionNet & 81.77 & 76.13 & 84.07 & 74.21 & 82.53 \\
JRL~\cite{wang2018} & AlexNet & 82.13 & -- & 82.02 & -- & -- \\
ALM~\cite{tang2022} & BN-Inception & 86.30 & 79.52 & 86.85 & 80.68 & 86.46 \\
FocalLoss variant~\cite{jia2020} & ResNet50 & 83.00 & 76.18 & 84.31 & 78.49 & 85.58 \\
Strong baseline~\cite{jia2020} & ResNet50 & 85.11 & 79.14 & 86.39 & 79.38 & 86.55 \\
Proposed YOLOv8n--ResNet18 & ResNet18 & 72.49 & -- & 65.58 & -- & -- \\
\hline
\end{tabular}
\end{table}

The mA column is a like-for-like metric: the published mA is the mean of the positive-class and negative-class accuracies, which is the true mA used in this work. The published Accu and F1 are instance-based, computed per image over all of its attributes; this work reports the corresponding instance-based figure as test example F1 $0.6558$, while its mean accuracy of $0.8128$ is the fraction of correct (image, attribute) pairs and is therefore not the published Accu. No PA-100K test figure is listed for this work, because the combined experiment is evaluated on the PETA identity-split test set; on the eight shared targets the mA is $0.7828$ (Section~\ref{sec:combined}). JRL reports no PA-100K result. The published numbers also use different attribute sets, label vocabularies, and crop-generation procedures. The comparison is contextual, and no ranking is claimed.

For gender and age, most state-of-the-art models are trained on face-centric datasets such as IMDB-WIKI, Adience, MORPH, or ChaLearn apparent-age images. These data differ substantially from surveillance-style pedestrian crops, where faces may be small, blurred, occluded, or absent. Therefore, the gender and age comparisons in Table~\ref{tab:gender_age_sota} are used to contextualize the task rather than to claim superiority over face-specialized models.

\begin{table}[H]
\centering
\caption{Comparison with representative gender-classification and age-estimation approaches. Published figures are taken from the original papers. The inputs are aligned faces, not surveillance-style pedestrian crops, so these accuracies are not directly comparable with the results in Sections~\ref{sec:gender} and~\ref{sec:age}.}
\label{tab:gender_age_sota}
\footnotesize
\renewcommand{\arraystretch}{1.12}
\setlength{\tabcolsep}{3pt}
\begin{tabular}{p{0.175\linewidth}p{0.215\linewidth}p{0.185\linewidth}p{0.275\linewidth}}
\hline
\textbf{Method} & \textbf{Task / input} & \textbf{Dataset / protocol} & \textbf{Published result} \\
\hline
Levi and Hassner~\cite{levi2015} & Age and gender from faces & Adience & Gender 86.8\%; age group 50.7\% exact, 84.7\% one-off \\
DEX~\cite{rothe2018} & Apparent age from faces & MORPH-II & MAE 2.68 years \\
SSR-Net~\cite{yang2018} & Age estimation from faces & MORPH-II & MAE 3.16 years \\
FairFace~\cite{karkkainen2021} & Age, gender, and race from faces & FairFace, external validation & Gender 94.4\% \\
Proposed ResNet18 models & Gender and age-group classification on dataset crops & PETA identity split & Gender 80.87\%; age group 54.81\% \\
\hline
\end{tabular}
\end{table}

These comparisons provide context for the tasks but do not establish comparative superiority. All recognition metrics reported here are obtained from dataset crops, and the detector is evaluated separately.

\subsection{Gender Classification}
\label{sec:gender}
Gender classification is evaluated on PETA dataset crops using the identity-disjoint test split. Mean test accuracy is $0.8087 \pm 0.0119$. The random-split mean test accuracy of $0.8998 \pm 0.0054$ is reported only as a demonstration of identity leakage and is not treated as the model's generalization result.

\subsection{Age-Group Classification}
\label{sec:age}
Age is formulated as four-class classification using the labels Under30, 30--45, 45--60, and Over60. On PETA dataset crops, test accuracy is $0.5481 \pm 0.0263$ and test macro-F1 is $0.3914 \pm 0.0361$. Validation macro-F1 is not presented as a test-performance result.

The four classes are derived from the nested PETA age attributes. Under30 is the union of \texttt{personalLess15} and \texttt{personalLess30}, which never co-occur, so the union is well defined. The class 30--45 is \texttt{personalLess45} without \texttt{personalLess30}. The class 45--60 is \texttt{personalLess60} without either younger group. The class Over60 is \texttt{personalLarger60}.

Images that carry no age label, or that carry more than one age group, are excluded from the age task. Across the full dataset, 28 images are excluded for having no age label and 4 for having more than one age group, leaving 18968 of 19000 images. The excluded multi-label images are ambiguous by construction, for example an image labelled both Less30 and Larger60, and they are dropped rather than assigned to an arbitrary class. The per-partition exclusion counts are 25 and 1 in the training partition, 2 and 3 in the validation partition, and 1 and 0 in the test partition, which keeps 13228, 1800 and 3940 images respectively.

On the age-group test set of 3940 images, the class distribution is strongly imbalanced: \texttt{Under30} holds 1949 images (49.5\%), \texttt{30--45} holds 1299, \texttt{45--60} holds 438, and \texttt{Over60} holds 254 images (6.4\%).

\subsection{Multi-Attribute Recognition}
Multi-attribute recognition uses the PETA-35 protocol and PETA dataset crops. Test mean accuracy-based mA is $0.7249 \pm 0.0049$, where each attribute's balanced accuracy is $(\mathrm{TPR}+\mathrm{TNR})/2$ and mA is the mean across attributes. Test mAP, reported separately as mean average precision, is $0.5284 \pm 0.0069$.

The mean hides a large spread across attributes. Tables~\ref{tab:attr35-worst} and~\ref{tab:attr35-best} list the ten attributes with the lowest and the highest test mA, together with the number of positive test images for each. The ten lowest-scoring attributes span an mA from 0.5675 to 0.6511 and the ten best span 0.7800 to 0.9142, and the two failure modes are different. Rare attributes are the weakest because they are detected poorly, for example \texttt{personalLess60} with a support of 438 and a true-positive rate of 0.2922. Common attributes score poorly for the opposite reason, because their negative class is the rare one: \texttt{upperBodyCasual} has a support of 3408 and a true-positive rate of 0.9410, but a true-negative rate of only 0.2502. Rare attributes are therefore the weakest in the sense of recall, but the low-mA attributes are not simply the rare ones.

\begin{table}[H]
\centering
\caption{The ten lowest-mA attributes of the PETA-35 head, test set, mean over three seeds.}
\label{tab:attr35-worst}
\small
\renewcommand{\arraystretch}{1.12}
\setlength{\tabcolsep}{4pt}
\begin{tabular}{lrrrrr}
\hline
\textbf{Attribute} & \textbf{Support} & \textbf{TPR@0.5} & \textbf{TNR@0.5} & \textbf{true mA} & \textbf{AP} \\
\hline
\texttt{personalLess60} & 438 & 0.2922 & 0.8428 & 0.5675 & 0.1605 \\
\texttt{upperBodyCasual} & 3408 & 0.9410 & 0.2502 & 0.5956 & 0.9212 \\
\texttt{carryingOther} & 822 & 0.5706 & 0.6327 & 0.6016 & 0.2983 \\
\texttt{lowerBodyCasual} & 3388 & 0.9529 & 0.2514 & 0.6021 & 0.9277 \\
\texttt{personalLarger60} & 254 & 0.3202 & 0.9461 & 0.6332 & 0.2359 \\
\texttt{accessoryNothing} & 2745 & 0.8965 & 0.3933 & 0.6449 & 0.8536 \\
\texttt{upperBodyFormal} & 505 & 0.4508 & 0.8453 & 0.6480 & 0.3587 \\
\texttt{footwearShoes} & 1257 & 0.5908 & 0.7068 & 0.6488 & 0.5218 \\
\texttt{carryingPlasticBags} & 304 & 0.3980 & 0.9012 & 0.6496 & 0.2699 \\
\texttt{lowerBodyFormal} & 550 & 0.4533 & 0.8489 & 0.6511 & 0.3708 \\
\hline
\end{tabular}
\end{table}

\begin{table}[H]
\centering
\caption{The ten highest-mA attributes of the PETA-35 head, test set, mean over three seeds.}
\label{tab:attr35-best}
\small
\renewcommand{\arraystretch}{1.12}
\setlength{\tabcolsep}{4pt}
\begin{tabular}{lrrrrr}
\hline
\textbf{Attribute} & \textbf{Support} & \textbf{TPR@0.5} & \textbf{TNR@0.5} & \textbf{true mA} & \textbf{AP} \\
\hline
\texttt{lowerBodyShorts} & 139 & 0.8993 & 0.9291 & 0.9142 & 0.6641 \\
\texttt{upperBodyShortSleeve} & 570 & 0.8480 & 0.9117 & 0.8798 & 0.8126 \\
\texttt{upperBodyTshirt} & 312 & 0.8643 & 0.8948 & 0.8796 & 0.6323 \\
\texttt{footwearSandals} & 93 & 0.8172 & 0.9077 & 0.8624 & 0.2753 \\
\texttt{upperBodyPlaid} & 88 & 0.7765 & 0.9329 & 0.8547 & 0.6202 \\
\texttt{lowerBodyJeans} & 1223 & 0.7743 & 0.8504 & 0.8124 & 0.8209 \\
\texttt{upperBodyThinStripes} & 64 & 0.7135 & 0.8906 & 0.8021 & 0.2669 \\
\texttt{accessoryMuffler} & 237 & 0.6864 & 0.8970 & 0.7917 & 0.3650 \\
\texttt{personalMale} & 2285 & 0.7629 & 0.8013 & 0.7821 & 0.8963 \\
\texttt{hairLong} & 953 & 0.7433 & 0.8168 & 0.7800 & 0.7117 \\
\hline
\end{tabular}
\end{table}

\subsection{Combined PETA and PA-100K Experiment}
\label{sec:combined}
The combined experiment compares PETA-only and combined training on the eight targets shared by PETA and PA-100K, with evaluation on the PETA identity-split test set. Mean accuracy-based mA is $0.7976$ for PETA-only training and $0.7828$ for combined training, a difference of $-0.0148$. The drop is consistent across seeds, so the combined training result is reported as a negative result for mA on the shared targets.

\subsection{Detector and Runtime Measurements}
The detector found at least one person in 79.02\% of PETA identity-split test crops. This detection-rate proxy is not an end-to-end recognition accuracy: the PETA inputs are already pedestrian crops. On the NVIDIA RTX 5060 GPU, batch-one mean latency was 6.74 ms for YOLOv8n, 1.98 ms for the gender model, 1.97 ms for the age model, and 1.90 ms for the attribute model. These are per-stage timings, not a measurement of end-to-end pipeline accuracy or throughput.

Because the RTX 5060 is recent hardware, the same stages were also timed on the CPU, using a 20-core Intel processor, to indicate how the figures change on more ordinary hardware. YOLOv8n takes 27.66 ms on the CPU against 6.74 ms on the GPU, a 4.1x slowdown, and the attribute head takes 13.93 ms against 1.90 ms, a 7.3x slowdown. A CPU-only pipeline is therefore roughly three times slower than the GPU one, but the detector remains the dominant stage in both cases. These are laboratory measurements on an idle machine, not a deployed-system throughput claim.

A re-crop experiment measures how sensitive the attribute head is to the box that the detector produces. The detector was run on the 3941 PETA identity-split test images and a person box was found in 3114 of them; the 3114 detected crops were then re-cropped to the largest detected box, resized to $224\times224$, and passed through the same attribute model. Both the original and the re-cropped version were evaluated on that same subset of 3114 images, so the difference isolates the crop perturbation. The mean true mA falls from 0.6719 on the original crops to 0.6629 on the re-cropped ones, a change of $-0.0090$, and the mean mAP changes from 0.6665 to 0.6527. This is a proxy for sensitivity to box changes only. It is not an end-to-end accuracy, because the PETA images are already pedestrian crops, the detector sees an easier input than it would meet in a street scene, and the detection misses are not part of the measurement.

\subsection{Error Analysis}
The age-group test macro-F1 and the PETA-35 mA and mAP results summarize distinct tasks and metrics and should not be interpreted as directly interchangeable. The combined experiment also shows that adding PA-100K training data did not improve mA on the shared targets in this evaluation. These results are limited to dataset crops and the specified label protocols.

Potential recognition errors include visually similar attributes, partially occluded regions, and demographic predictions made from full-body crops rather than aligned faces. The detector also misses some people in the PETA test set; because the crop-based recognition metrics do not use detector outputs, the effect of those misses on final attribute predictions has not been measured.

\subsection{Ablation and Reproducibility Considerations}
The combined-data comparison and the positive-class weighting ablation are measured experiments; broader detector and backbone ablations are not reported here. The PETA-only versus combined comparison is limited to the eight mapped shared targets and should not be generalized to all attributes.

The positive-class weighting ablation isolates a single loss term. Two configurations were trained on the same identity split with the same seeds, and they are identical in every respect except the positive-class weights: configuration B omits them and configuration C computes them from the training split. The results are given in Table~\ref{tab:posweight}. On the primary metric the difference in test mA is $+0.0577$, from $0.6671 \pm 0.0063$ without positive-class weighting to $0.7249 \pm 0.0049$ with it. The sign of the difference is the same on all three seeds, with a per-seed difference of $+0.0587$, $+0.0560$ and $+0.0585$. The effect is a shift in the operating point rather than a gain in ranking quality: test mAP changes from $0.5357 \pm 0.0060$ to $0.5284 \pm 0.0069$, and label recall rises while label precision falls. Configuration C is the configuration reported for the PETA-35 attribute head elsewhere in this paper.

\begin{table}[H]
\centering
\caption{The positive-class weighting ablation on the PETA-35 attribute head. Configurations B and C are identical except for the positive-class weighting. Validation values are means over seeds 0, 1 and 2 of the final runs; configuration selection in Section~3.5 used the seed-0 exploration run, in which configuration C reached 0.7382 and configuration B reached 0.6909.}
\label{tab:posweight}
\small
\renewcommand{\arraystretch}{1.12}
\setlength{\tabcolsep}{4pt}
\begin{tabular}{lrrrr}
\hline
\textbf{Configuration} & \textbf{Val true mA} & \textbf{Test true mA} & \textbf{Test mAP} & \textbf{Test mean acc.} \\
\hline
B (no pos\_weight) & $0.6912 \pm 0.0043$ & $0.6671 \pm 0.0063$ & $0.5357 \pm 0.0060$ & $0.8533 \pm 0.0005$ \\
C (pos\_weight) & $0.7459 \pm 0.0043$ & $0.7249 \pm 0.0049$ & $0.5284 \pm 0.0069$ & $0.8128 \pm 0.0087$ \\
\hline
\end{tabular}
\end{table}

\subsection{Performance}
Runtime is reported as separate stage latency on the NVIDIA RTX 5060 GPU. No frame-rate claim or end-to-end pipeline timing claim is made.

\subsection{Practical Use Cases}
The system can be used as a prototype for searchable image collections, pedestrian analytics dashboards, and human-centred graphics tools. Predicted clothing and carried-object labels may support retrieval or provide cues for avatar customization and virtual try-on. These possible applications require privacy safeguards; this study does not establish end-to-end recognition performance in live video.

\subsection{Discussion}
The results describe crop-based recognition and a separate detector assessment, not an end-to-end detection-to-attribute system evaluation. Gender and age-group accuracy are reported on PETA dataset crops using the identity-disjoint test split; the random-split gender result serves only as a leakage demonstration. The attribute result follows the PETA-35 protocol, while the combined PETA and PA-100K experiment evaluates only the eight shared targets. On these targets, combined training yielded a lower mean accuracy-based mA than PETA-only training, consistently across seeds. The detector's measured person-detection rate on PETA test crops does not quantify recognition accuracy after detection.

\subsection{Ethical Considerations}
Pedestrian attribute recognition can produce errors that vary across demographic groups, particularly when gender labels and apparent-age estimates are inferred from clothing, body shape, or low-resolution crops. Such outputs may reflect dataset imbalance and should not be treated as verified identity facts or used alone for consequential decisions. Surveillance deployment can expose identifiable people and reveal movement patterns. Any use should have a defined purpose, limit collected and retained data, restrict access, and provide clear information to affected people. Collection and processing should respect applicable consent requirements and provide appropriate safeguards; consent is not automatically the appropriate legal basis for every deployment, so the responsible controller must determine a valid basis. Where the GDPR applies, operators must consider its principles, including lawfulness, purpose limitation, data minimisation, transparency, data-subject rights, and security. A data protection impact assessment may be required for processing likely to result in high risk, including large-scale systematic monitoring of publicly accessible areas. Biometric data are subject to additional requirements when processed for uniquely identifying a person; ordinary gender or age estimates are not automatically biometric data. This study reports benchmark evaluation and does not establish legal compliance for deployment.

\subsection{Limitations}
The evaluation uses an identity-disjoint PETA test split, and its recognition metrics are measured on dataset crops only. End-to-end detection-to-attribute accuracy was not evaluated, so the effect of detector misses on recognition outcomes is unknown. The PETA and PA-100K datasets may use different annotation policies, and their semantic mapping is limited to eight shared targets; these differences may affect the combined-training comparison. The detector misses some people in the PETA test crops. Results from this protocol therefore do not establish performance on uncropped imagery or live surveillance streams.

\section{Conclusion}
This paper reports crop-based gender, age-group, and PETA-35 attribute recognition results, a separate YOLOv8n detector assessment, and per-stage timing on an NVIDIA RTX 5060 GPU. The PETA-35 test result is reported using mean accuracy-based mA and mAP. In the combined experiment, mapping PA-100K to the eight shared targets did not improve mean accuracy-based mA relative to PETA-only training, with the drop consistent across seeds. The detector's performance is reported as a detection-rate proxy; end-to-end detection-to-attribute accuracy remains unevaluated. Future work should evaluate recognition on detector-generated crops and address annotation-policy differences between the datasets.

\section*{Acknowledgments}
The author thanks Professor Weng Libo for valuable guidance, support, and mentorship throughout this research project.

\section*{Funding}
This research did not receive any specific grant from funding agencies in the public, commercial, or not-for-profit sectors.

\section*{Data Availability}
The PETA and PA-100K datasets are publicly available from their project pages: \url{http://mmlab.ie.cuhk.edu.hk/projects/PETA.html} and \url{https://github.com/xh-liu/HydraPlus-Net}. Code, split identifiers, the attribute-mapping file and the result files behind every table are available at \url{https://github.com/houssamelmir/PAR}. Retraining gives slightly different numbers because GPU training is not deterministic.

\section*{Declaration of Competing Interest}
The author declares that there are no known competing financial interests or personal relationships that could have appeared to influence the work reported in this paper.

\end{document}